\documentclass[hidelinks]{new_tlp}

\usepackage{mathptmx}
\usepackage{graphicx}
\usepackage{subcaption}
\usepackage{amsmath}
\usepackage{hyperref}
\usepackage{colortab}
 
\newcommand{\cellcontent}[1]{``#1''}
\newcommand{\specification}[1]{``#1''}
\newcommand{\cdmnobj}[1]{\textit{#1}}
\newcommand{\cdmnvar}[1]{\texttt{#1}}

\newcommand{\chalname}[1]{\textit{#1}}
\newcommand{\bigforall}{\mathop{\mbox{\Large $\forall$}}}
\newcommand{\bigexists}{\mathop{\mbox{\Large $\exists$}}}

\newcommand{\fodot}[0]{FO($\cdot$)}

\newcommand{\Color}[1]{%
        \pdfcolorstack\pdfcolorstackinit page direct{0 g}%
             push {#1 rg}%
         }
\newcommand{\graycell}{\Color{0.8 0.8 0.8}}
\newcommand{\resetcolor}{\Color{0 0 0}}

\newcommand{\new}[1]{{#1}}
\newcommand{\old}[1]{}

\usepackage[normalem]{ulem}

\title[Constraint Decision Model and Notation]
      {Tackling the DM Challenges with cDMN:\\
       A Tight Integration of DMN and Constraint Reasoning}

\author[S. Vandevelde, B. Aerts, J. Vennekens]
       {SIMON VANDEVELDE\thanks{This research received funding from the Flemish Government under the “Onderzoeksprogramma Artificiële Intelligentie (AI) Vlaanderen” programme.}, BRAM AERTS, JOOST VENNEKENS\\
       KU Leuven, De Nayer Campus, Dept. of Computer Science\\
       J.-P. De Nayerlaan 5, 2860 Sint-Katelijne-Waver, Belgium\\
       Leuven.AI - KU Leuven Institute for AI, B-3000 Leuven, Belgium\\
       \email{\{s.vandevelde, b.aerts, joost.vennekens\}@kuleuven.be}}

\jdate{March 2003}
\pubyear{2003}
\pagerange{\pageref{firstpage}--\pageref{lastpage}}
\doi{}

\begin{document}

\label{firstpage}

\maketitle

  \begin{abstract}
    Knowledge-based AI typically depends on a knowledge engineer to construct a formal model of domain knowledge --- but what if domain experts could do this themselves? 
    This paper describes an extension to the Decision Model and Notation (DMN) standard, called Constraint Decision Model and Notation (cDMN). 
    DMN is a user-friendly, table-based notation for decision logic, which allows domain experts to model simple decision procedures without the help of IT staff.
    cDMN aims to enlarge the expressiveness of DMN in order to model more complex domain knowledge, while retaining DMN’s goal of being understandable by domain experts. 
    We test cDMN by solving the most complex challenges posted on the DM Community website. 
    We compare our own cDMN solutions to the solutions that have been submitted to the website and find that our approach is competitive\old{, both in readability and compactness}. 
    Moreover, cDMN is able to solve more challenges than any other approach.

    Under consideration in Theory and Practice of Logic Programming (TPLP).
  \end{abstract}

  \begin{keywords}
    Decision Model and Notation, constraint reasoning, expressiveness, readability, IDP system
  \end{keywords}

\tableofcontents

\section{Introduction}

The Decision Model and Notation (DMN) \cite{DMN} standard, designed by the Object Management Group (OMG), is a way of representing data and decision logic in a \old{readable,} table-based way.
It is intended to be used directly by business experts without the help of computer scientists \new{, and as such, aims to be low in complexity and user-friendly}.

While DMN is very effective in modelling deterministic decision processes, it lacks the ability to represent more complex kinds of knowledge.
In order to explore the boundaries of DMN, the Decision Management Community website\footnote{https://dmcommunity.org/} issues a monthly decision modelling challenge.
Community members can then submit a solution, using their preferred decision modelling tools or programming languages.
This allows solutions for complex problems to be found and compared across multiple DMN-like representations.
So far, none of the available solvers have been able to solve all challenges. 
Moreover, the available solutions sometimes fail to meet the readability goals of DMN, because the representation is either too complex, too large or requires a specific computer science background.

In this paper, we propose an extension to the DMN standard, called cDMN.
It aims to allow more complex knowledge to be represented, while remaining readable by business users. 
The main features of cDMN are constraint modelling, quantification, and the use of concepts such as types and functions.
We test the expressiveness of cDMN  on the decision modelling challenges.

In \citeN{Deryck}, we presented a preliminary framework for constraint modelling in DMN\@. In the current paper, we extend this by adding quantification, types, functions, relations, data tables, optimization and by evaluating the resulting cDMN formalism on the DMN challenges.

This paper is an extended version of a paper we presented at the RuleML+RR 2020 conference \cite{cDMN}. 
It includes an updated list of challenges, changes to the semantics to make it more complete, a more in-depth description of the solver and a section on the integration of DMN into business models.

It is structured as follows. 
In Section~\ref{sec:prelim} we briefly describe the DMN standard.
Section~\ref{sec:chaloverview} gives an overview of the challenges used in this paper. 
After this, we touch on the related work in Section~\ref{sec:relatedwork}.
We discuss both syntax and semantics of our new notation in Section~\ref{sec:cDMN}.
Section~\ref{sec:implementation} briefly discusses the implementation of our cDMN solver.
We compare our notation with other notations and evaluate its added value in Section~\ref{sec:results}, and conclude in Section~\ref{sec:conclusion}.

\section{Preliminaries: DMN}\label{sec:prelim}

The DMN standard \cite{DMN} describes the structure of a DMN model.
\new{
The aim of the standard is to provide a user-friendly modelling notation for decision logic.
It is suitable for use by business experts \cite{silver2018}, low in complexity \cite{complexity}, and has already been successfully used in many case studies \cite{Sooter2019,Car2018,HasicUC2020}.
}

\old{Such a model}\new{A DMN model} consists of two components: a Decision Requirements Diagram (DRD), and a number of decision tables. 
The DRD is a graph that expresses the structure of a DMN model by representing the connections between inputs, decisions, knowledge sources and more.
At its core, it is a visual representation of \old{what information is needed to make a decision} \new{the general structure of the model, depicting which concepts are defined in terms of which other concepts}.
\new{As such, it improves the interpretability of models by end-users.} 
\old{This improves the interpretability of models by end-users.}
Furthermore, it also enhances the traceability of decisions, as it becomes easier to \old{check why a certain decision was made \new{by a DMN execution engine}} \new{see which variables influence a specific decision, and in which order different decisions depend on each other}.
\old{Indeed, by starting at the base inputs, a user can work their way up the graph to the final decision(s) and manually verify a result\new{, analogously to the engine}.}

\begin{figure}
    \includegraphics[width=5cm]{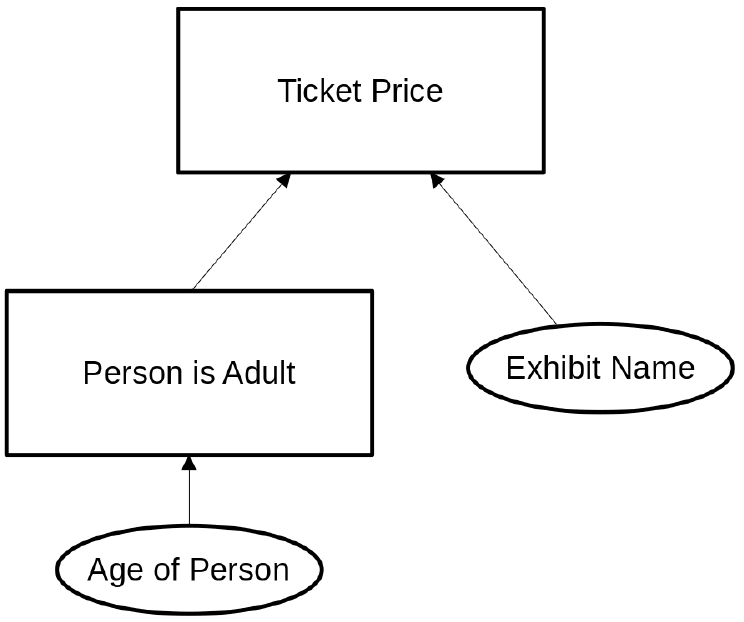}
    \caption{Decision Requirement Diagram to decide a museum ticket price in DMN.}
    \label{fig:drd}
\end{figure}

An example of a DRD is shown in Figure~\ref{fig:drd}, which represents the decision process for the cost of an entry ticket for a museum.
The two ellipses represent input data, i.e., a person's age and the name of the exhibit they want to visit.
Each rectangle in the graph represents a decision that needs to be made, e.g.\new{,} deciding whether a person is an adult.

\begin{figure}
    \includegraphics[width=\linewidth]{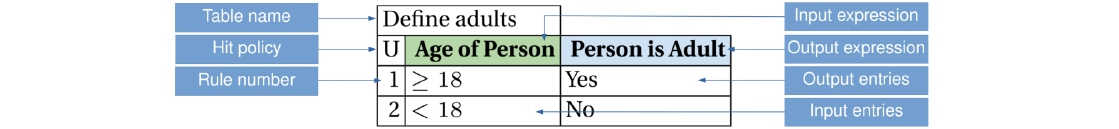}
    \caption{Decision table to define whether a person is an adult.}
    \label{fig:decisiontable}
\end{figure}

The decision tables contain the in-depth business logic of a model.
An example of such a decision table can be found in Figure \ref{fig:decisiontable}. 
It consists of a number of input columns (darker green) and \old{a single} \new{at least one} output column (lighter blue). 
Each row is read as: if the input conditions are met (e.g., if \cellcontent{Age of Person} satisfies the comparison \cellcontent{$\geq$ 18} ), then the output variable is assigned the value of the output entry (e.g. \cellcontent{Person is Adult} is assigned value \cellcontent{Yes}). 
Only single values, such as strings and numbers, can be used as output entries.
In the case where no row matches the input, then each output is either set to the special value \textit{null} (which is typically taken to indicate an error in the specification) or to the output's default value, if one was provided.

The behaviour of a decision table is determined by its hit policy.
There are a number of \emph{single hit} policies, which \old{define that a table can have at most one output \new{value} for each possible input, such as} \new{cause the output variable(s) to take on a single value, even when multiple rows are applicable. In particular, these hit policies are as follows:} \cellcontent{Unique} (for each possible set of input values, at most one row is applicable), \cellcontent{Any} (if multiple rows are applicable for the same input values, their outputs must be the same) and \cellcontent{First} (if more than one row is applicable, the first applicable row determines the value of the outputs). There exist also \emph{multiple hit} policies such as \cellcontent{Collect} (collect the output of all applicable rows in a list) and \cellcontent{C+} (sum the output of all applicable rows).
Regardless of which hit policy is used, each decision table uniquely determines the value of its output(s).

The entries in a decision table are typically written in the (Simple) Friendly Enough Expression Language, or (S-)FEEL, which is also part of the DMN standard. 
S-FEEL allows simple values, lists of values, numerical comparisons, ranges of values and arithmetic expressions.
Decision tables with S-FEEL are generally considered quite readable by domain experts.

In addition, DMN also allows more complex FEEL statements in combination with boxed expressions, as will be illustrated in Figure \ref{fig:FEELmap}.
However, this also greatly increases complexity of the representation, which makes it unsuitable for use by domain experts without the aid of knowledge engineers.

\section{Challenges Overview}\label{sec:chaloverview}

Of all the challenges on the DM Community website, we selected those that did not have a straightforward DMN-like solution.
The list of the 24 challenges that meet this criterion can be found in Table \ref{table:challenges_properties}.
We categorize these challenges according to four different properties.
Table \ref{table:problemproperties} shows the list of properties, and the percentage of challenges that have this property.

The most frequent property is the need for aggregates (54.17\%), such as counting the number of violated constraints in \chalname{Map Coloring with Violations} or summing the number of calories of ingredients in \chalname{Make a Good Burger}. 
The second most frequent property is having constraints in the problem description (37.50\%). 
For instance, the constraint in \chalname{Map Coloring} states that two bordering countries can not share the same color.
The next property, universal quantification (33.33\%), is that a statement applies to every element of a type, for example in \chalname{Who Killed Agatha?}: nobody hates everyone.
The final property, optimization, occurs in 25.00\% of the challenges.
For example, in \chalname{Zoo, Buses and Kids} the cheapest set of buses must be found.

\new{
The description of each challenge can be found on the DMCommunity website\footnote{\url{https://dmcommunity.org/challenge/}}, together with their submissions.
We also maintain a mirror repository\footnote{\url{https://gitlab.com/EAVISE/cdmn/DMChallenges}} containing the specific challenges and submissions used in this work.
}

\begin{table}
    \centering
    \caption{List of DM Community challenges and their properties. 1: Universal Quantification, 2: Constraints, 3: Optimization, 4: Need for Aggregates.}
    \label{table:challenges_properties}
    \begin{tabular}{c r c c r}
        \hline \hline
        Challenge & Property & & Challenge & Property\\
        \hline
        Who Killed A.? & 1 & & Change Making & 3, 4\\
        A Good Burger & 2, 3, 4 & & Define Dupl. & None\\
        Coll. of Cars & None & & Monkey Business & None\\
        Vacation Days & 1, 2, 4 & & Family Riddle & 2, 4\\
        Cust. Greeting & None & & Online Dating & None\\
        Loan Approval & 4 & & Class. Employees & 4\\
        Soldier Payment & 4 & & Reinder Order & None\\
        Zoo, Buses, Kids & 3, 4 & & Balanced Assign. & 3\\
        Vac. Days Adv. & 1, 2, 4 & & Map Coloring & 1, 2\\
        Map Color Viol. & 1, 2, 3, 4 & & Crack The Code & 4\\
        Numerical Haiku & 1, 2, 4 & & Nim Rules & 2\\
        Doctor Planning & 1, 2, 4 & & Calculator & 1, 3\\
        \hline \hline
    \end{tabular}
\end{table}

\begin{table}
    \caption{Percentage of occurrence of properties in challenges.}
    \label{table:problemproperties}
    \begin{tabular}{l r}
        \hline\hline
        Property & (\%) \\
        \hline
        1. Aggregates needed        & 54.17  \\
        2. Constraints              & 37.50  \\
        3. Universal quantification & 33.33  \\
        4. Optimization             & 25.00  \\
        \hline\hline
    \end{tabular}

\end{table}

\section{Related Work}\label{sec:relatedwork}

It has been recognized that even though DMN has many advantages, it is somewhat limited in expressiveness \cite{Calvanese2019,Deryck}.
This holds especially for decision tables with S-FEEL, the fragment of FEEL that is considered most readable.
While full FEEL is more expressive, it is not suitable to be used by domain experts without the aid of knowledge engineers.
Moreover, it does not provide a solution to other shortcomings, such as the lack of constraint reasoning and optimization.

One of the systems that does effectively support constraint solving in a readable DMN-like representation is the OpenRules system \cite{OpenRules}. 
It enables modellers to define constraints over the solution space by allowing \textit{Solver Tables} to be added alongside decision tables. 
In contrast to standard decisions, which assign a specific value to an output, Solver Tables allow for setting constraints on the output space.
OpenRules offers a number of \textit{DecisionTableSolve-Templates}, which can be used to specify these constraints.
It is possible to either use these predefined templates, or to define such a template manually if the predefined ones are not expressive enough. 
Even though this system extends the range of applications that can be handled, there are three reasons why it does not offer the ease of use for business users that we are after. 
First, because of the wide range of available templates for solver tables, which differ from that of standard decision tables, using the OpenRules constraint solver entails a steep learning curve.
Second, the solver's functionality can only be accessed through the Java API, which goes against the DMN philosophy \cite[p.~13]{DMN}.
Third, because of the lack of quantification in OpenRules, solutions are generally not independent of domain size, which reduces readability.

Another system that aims to increase expressiveness of DMN is Corticon \cite{Corticon}. 
It implements a basic form of constraint solving by allowing the modeller to filter the solution space.
While this approach indeed improves expressiveness, it decreases readability.
Moreover, some constraints can only be expressed by combining a number of rules and a number of filters. 
For example, when expressing \specification{all female monkeys are older than 10 years}, this is split up in two parts; (1) a rule that states that if \texttt{Monkey.gender = female \& Monkey.Age < 10 THEN Monkey.illegal = True} and (2) a filter that states that a monkey cannot be illegal: \texttt{Monkey.illegal = False}.
There are no clear guidelines about which part of the constraints should be in the filter and what should be a rule.
A more detailed comparison between OpenRules, Corticon and cDMN is given in Section \ref{sec:results}.

\citeN{Calvanese2019} propose an extension to DMN which allows for expressing additional domain knowledge in Description Logic\new{, which would not be possible to model in DMN}.
\new{In this way, t}\old{T}hey share our goal of extending DMN to express more complex real-life problems.
However, they introduce a completely separate Description Logic formalism, which \old{seems} \new{may be} too complex for a domain expert to use.
\new{While this approach makes sense if, e.g., a Description Logic ontology for the domain is already available, it seems less suited for cases in which a domain expert would need to construct this.}
Unfortunately, they did not submit any solutions to the DMN Challenges, which leaves us unable to compare its expressiveness in practice.

\section{cDMN: Syntax \& Semantics}\label{sec:cDMN}

While DMN allows modellers to elegantly represent a deterministic decision process, it lacks the ability to specify constraints on the solution space.
The cDMN framework extends DMN, by allowing constraints to be represented in a straightforward \old{and readable} manner.
It also allows for representations that are independent of domain size by supporting types, functions, relations and quantification.
To select one or more solutions from the solution space, multiple inferences tasks are supported.

We now explain both the usage and the syntax of every kind of table present in cDMN.

\subsection{Glossary}\label{sub:glossary}

\begin{figure}
    \centering
    \includegraphics[width=\linewidth]{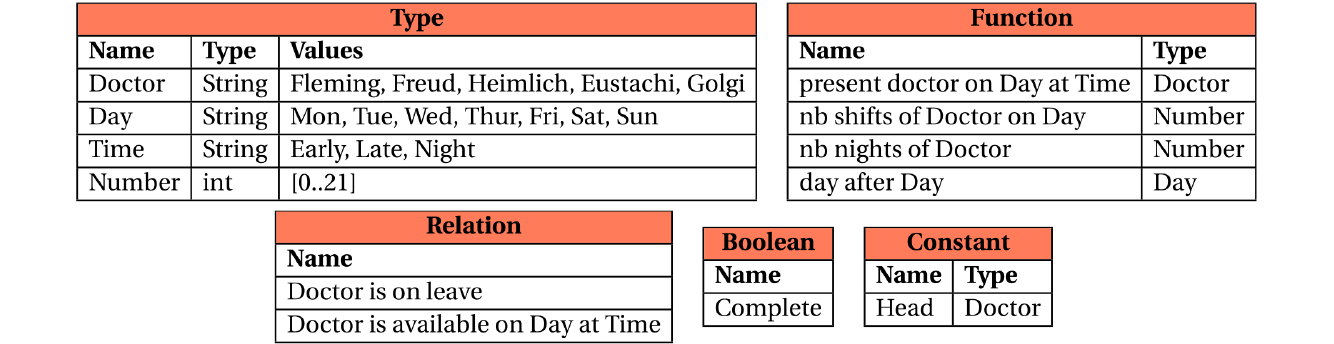}
    \caption{An example cDMN glossary for the \chalname{Doctor Planning} problem.}
    \label{tab:glossary_example}
\end{figure}

In logical terms, the ``variables'' of standard DMN correspond to constants (i.e., 0-ary functions).
cDMN extends these by adding $n$-ary functions and $n$-ary relations. 
Similarly to OpenRules and Corticon, we allow the modeller to define their vocabulary by means of a glossary.
It consists of at most five glossary tables, each enumerating a different kind of symbol.
An example glossary for the \chalname{Doctor Planning} challenge is given in Figure \ref{tab:glossary_example}.

In the \cdmnobj{Type} table, \cdmnobj{type} symbols are declared. 
The value of each type is a set of domain elements, specified either in the glossary or in a data table (see Section \ref{sec:datatable}).
An example is the type \cdmnvar{Doctor}, which contains the names of doctors.
\new{By convention, type symbols start with a capital letter.}

In the \cdmnvar{Function} table, a symbol can be declared as a \textit{function} of one or more types to another. 
\new{There is no fixed syntax for functions; all types that appear in the description are interpreted as arguments to the function (of this type) and the remaining text is the name of the function.}
\old{The argument(s) of a function are those types that appear in its name, and they are mapped on the type listed in the \cdmnobj{Type} column.}
For example, \old{the} \cdmnvar{nb nights of Doctor} \new{has one argument of type \cdmnvar{Doctor}, and ``\textit{nb nights of}'' is its name.
Intuitively, this } function denotes how many nights a doctor works per week.
It maps each element of type \cdmnvar{Doctor} to an element of type \cdmnvar{Number}. 
Functions with $n > 1$ can be defined by using $n$ arguments in the name, such as \cdmnvar{present doctor on Day at Time}, which assigns a doctor to every pair of \cdmnvar{Day} and \cdmnvar{Time}.
\new{
The detection of arguments is case sensitive, so \cdmnvar{doctor} is not considered an argument, but \cdmnvar{Doctor} is.
}

For each domain element, a constant with the same name is automatically introduced, which allows the modeller to refer to this domain element in constraint or decision tables.
For instance, the modeller can use the constant \cdmnvar{Fleming} to refer to the domain element \cdmnvar{Fleming}.
In addition, the \cdmnobj{Constant} table allows also other constants to be introduced.
Recall that such logical constants correspond to standard DMN variables.
In our example case, we use a constant \cdmnvar{Head} of type \cdmnvar{Doctor}, which means it can refer to any of the domain elements \cdmnvar{Fleming}, \cdmnvar{Freud}, \cdmnvar{Heimlich}, \cdmnvar{Eustachi} or \cdmnvar{Golgi}.

In the \cdmnobj{Relation} table, a verb phrase can be declared as a \textit{relation} on one or more given types.
For instance, the relation \cdmnvar{Doctor is on leave} denotes for each \cdmnvar{Doctor} whether they are on leave.
Similarly to functions, \new{there is no strict syntax:} $n$-ary predicates can be defined by using $n$ arguments in the name, e.g. \cdmnvar{Doctor is available on Day at Time} is a relation with three arguments (respectively of the type \cdmnvar{Doctor}, \cdmnvar{Day} and \cdmnvar{Time}), that denotes whether a doctor is available on a specific day, at a specific time.

The \textit{Boolean} table contains \textit{boolean} symbols (i.e., propositions), which are either true or false.
An example is the boolean \texttt{Complete}, which denotes whether the planning is complete.

\subsection{Decision Tables and Constraint Tables}\label{sub:decision_constraint}

As stated earlier in Section \ref{sec:prelim}, a standard decision table uniquely defines the value of its outputs.
We extend DMN by allowing a new kind of table, called a \textit{constraint table}, which does not have this property.

Whereas decision tables only allow single values to appear in output columns, our constraint tables allow arbitrary S-FEEL expressions in output columns.
Each row of a constraint table represents a logical \textit{implication}, in the sense that, if the conditions on the inputs are satisfied, then the conditions on the outputs must also be satisfied.
This means that if, for instance, none of the rows are applicable, the outputs can take on an arbitrary value, as opposed to being forced to \textit{null}.
In constraint tables, no default values can be assigned.
Because of these changes, a set of cDMN tables does not define a single solution, but rather a solution space containing a set of possible solutions.

\begin{figure}
    \centering
    \includegraphics[width=\linewidth]{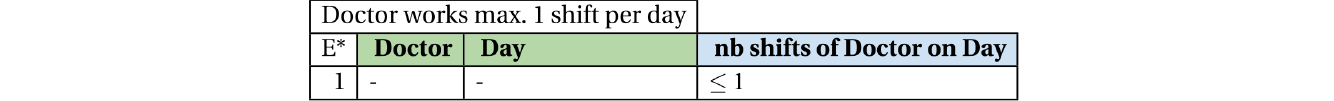}
    \caption{Constraint table to express that a doctor works a maximum of one shift per day.}
    \label{fig:maxshift}
\end{figure}

We introduce a new \textit{hit policy} to identify constraint tables. We call this the \cellcontent{Every} hit policy, denoted as \cellcontent{E*}, because it expresses that every implication in the table must be satisfied. 
An example of this can be found in Figure~\ref{fig:maxshift}, which states that every doctor can work a maximum of one shift per day.

cDMN does not only introduce constraint tables, it also extends the expressions that are allowed in column headers, both in decision and constraint tables.
There are two types of headers in cDMN: the \textit{term-denoting} headers, and the \textit{atom-denoting} headers.
A term-denoting header can consist of the following five expressions.
\begin{enumerate}
    \item  A type $\mathit{Type}$. Such expression introduces a new variable $\mathit{x}$ of type $\mathit{Type}$, which is only defined in the scope of the table.
    \item  An expression of the form ``$\mathit{Type}$ called $\mathit{name}$''. This expression introduces a new variable $\mathit{name}$ of the type $\mathit{Type}$ in the scope of the table.
    \item  A constant.
    \item  An arithmetic combination of term-denoting header expressions (such as a sum of constants).
    \item  A function expression such as ``$\mathit{Function}$ of $\mathit{arg}_1$ and \ldots and $\mathit{arg}_n$'', where each of the $\mathit{arg}_i$ is a term-denoting header expression, or a previously introduced variable. This expression applies the function to its arguments.
\end{enumerate}

An atom-denoting header consists of a relation expression such as ``$\mathit{Relation}$ for $\mathit{arg}_1$ and \ldots and $\mathit{arg}_n$'', where each of the $\mathit{arg}_i$ is a term-denoting header expression, or a previously introduced variable.
This expression applies the relation to its arguments.

The first two kinds of term-denoting expressions are called {\em variable} header expressions. 
They allow \textit{universal quantification} in cDMN.
Each input column whose header consists of such a \emph{variable} expression either introduces a new universally quantified variable (we call this a \emph{variable-introducing} column), or refers back to a variable introduced in a preceding variable-introducing column.
Once a variable $x$ has been introduced by an expression \textit{Type} (item 1), subsequent uses of the expression \textit{Type} refer back to this variable $x$.
Similarly, once a named variable \textit{name} has been introduced by an expression \textit{Type} called \textit{name} (item 2), subsequent uses of the expression \textit{name} refer back to this variable \textit{name}.

\begin{figure}
    \centering
    \includegraphics[width=\linewidth]{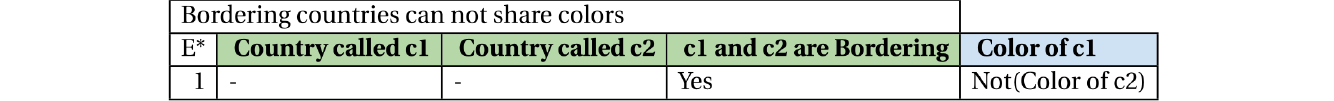}
    \caption{Example of a constraint table with quantification in cDMN, defining that bordering countries can not share colors.}
    \label{tab:map}
\end{figure}

The table in Figure~\ref{fig:maxshift} shows an example of quantification in cDMN. 
It introduces universally quantified variables of the type \textit{Doctor} and \textit{Day}, places no restrictions on these variables (i.e. ``-''), and hence states that every doctor can only work a maximum of one shift on every day.
To illustrate the use of named variables, Figure~\ref{tab:map} defines variables \texttt{c1} and \texttt{c2}, both of the type \texttt{Country}, and states that when those countries are bordering, they cannot have the same color.

In summary, this section has discussed three ways in which cDMN extends DMN.
First, the hit policy \cellcontent{E*} changes the semantics of the table from a definition to a set of implications.
Second, constraint tables allow S-FEEL expressions in the output columns.
Third, cDMN allows quantification, functions, predicates to be used in both decision tables and constraint tables.

\subsection{Data Tables} \label{sec:datatable}

\begin{figure}
    \centering
    \includegraphics[width=\linewidth]{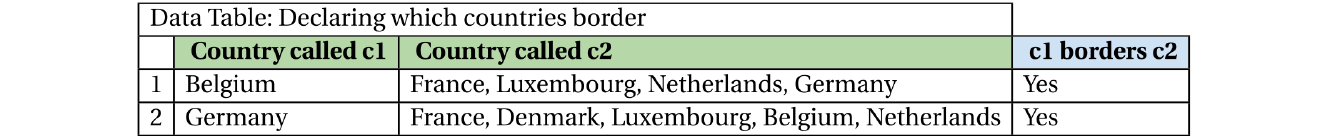}
    \caption{Data table describing countries and their neighbours}
    \label{tab:datatable_example}
\end{figure}

Typically, problems can be split up into two parts: (1) the general logic of the problem, and (2) the specific problem instance that needs to be solved. 
Take for example the map coloring problem: the general logic consists of the rule that two bordering countries cannot share a color, whereas the instance of the problem is the specific map (e.g., Western Europe) to color.
cDMN extends the DMN standard to include \cdmnobj{data tables}, which are used to represent the problem instances, separating them from the general logic.
The format of a data table closely resembles that of a  decision table, with a couple of exceptions.
Instead of a hit policy, a data table has ``data table'' in its name.
Furthermore, only basic values (integers, floats and elements of a type) are allowed in data tables.
It is also possible for columns to have more than one value in a certain cell, in which case the row is instantiated for each of these values.
Since functions in cDMN models are always assumed to be total, a data table for a function should be complete, i.e., there should be a value defined for every possible combination of input arguments.
As an example, a snippet of the data table for the \chalname{Map Coloring} challenge is shown in Figure \ref{tab:datatable_example}.

Data tables offer several advantages.

\begin{enumerate}
    \item There is a methodological advantage: by separating data tables from decision tables, it becomes easier to reuse the specification.
    \item If the modeller chooses to enumerate the domain of a type in the glossary, then the system checks that each value in a data table indeed belongs to the domain of the appropriate type. 
    This helps to prevent errors or typos in the input data or glossary.
    If the modeller chooses not to enumerate a type in the glossary, then the type's domain defaults to the set of all values in the data table. 
    \item The cDMN solver is able to compute solutions faster, due to a different internal representation between data tables and decision tables.
\end{enumerate}

\subsection{Goal Table}

A standard DMN model defines a deterministic decision procedure.
It is typically always used in the same way: the external inputs are supplied by the user, after which the values of the output variables are computed by forward propagation.

\begin{figure}
    \centering
    \includegraphics[width=\linewidth]{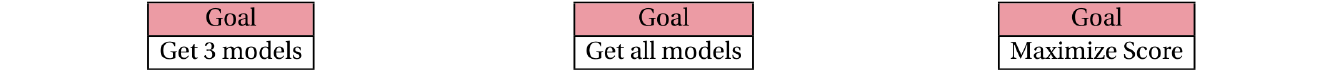}
    \caption{Goal table examples}
    \label{fig:execute}
\end{figure}

When using the cDMN solver, this is no longer the case.
We can fill in as many or as few variables as we want, and use the cDMN specification to derive useful information about the not-yet-known variables.
By employing a \textit{goal} table, modellers can state what the specification is to be used for: model expansion or optimization.
Model expansion is the task of finding an interpretation for each of the symbols (a ``model'', in the terminology of classical logic) that satisfies all of the tables, and optimization is the task of finding the model with either the lowest or highest value for a given term.
Examples of such tables are given in Figure~\ref{fig:execute}.

In summary, a cDMN model consists of:
\begin{itemize}
    \item A glossary;
    \item A set of data tables;
    \item A set of constraint tables;
    \item A set of decision tables;
    \item At most one goal table.
\end{itemize}
Apart from the glossary, all other kinds of tables are optional.

\subsection{Semantics of cDMN}\label{ss:semantics}

The meaning of a cDMN specification is given by a possible world semantics.
As in classical logic, a possible world is represented by a structure $S$ for a vocabulary $V$.
Such a structure consists of a domain $D$ and an assignment of each symbol $\sigma \in V$ to an appropriate relation/function $\sigma^S$ on $D$.
We will define the semantics of cDMN by means of a translation to \fodot, which is a typed variant of classical FO that also extends it with a number of additional constructs such as aggregates \cite{BruynoogheMaurice2015TPLP,DeCatBroes2018PLaa,wittocx2008idp}.
In this typed logic, a number of unary predicates are designated as types and each structure $S$ must be such that the interpretations $t_i^S$ of the types $t_i$ form a partition of the domain of $S$.
In addition, each relation/function $\sigma$ has a typing, which must be respected by the interpretation $\sigma^S$, i.e., if a predicate $P$ has typing $(T_1, \ldots, T_n)$ then $P^S \subseteq T_1 \times \ldots \times T_n$.

We will define the set of possible worlds for a cDMN model as follows.
The DMN glossary defines a typed \fodot{} vocabulary $V$ in a straightforward way.
The data tables, together with the glossary, define a structure $S$ for a part $V' \subseteq V$ of this vocabulary: i.e., the domain of $S$ is defined, as well as the interpretation $\sigma^S$ of the symbols $\sigma \in V'$; however, for the remaining symbols $\sigma \in V' \setminus V$, the data tables do not yet define an interpretation.
We will translate the decision and constraint tables into a theory $T$ of \fodot{} sentences, such that the possible worlds of a cDMN model are precisely the structures $S'$ that extend $S$ with an interpretation for the remaining symbols $V' \setminus V$ in such a way that $S' \models T$, i.e., that all the decision/constraint tables are satisfied.

What remains is to transform each of the decision and constraint tables into an \fodot{} sentence.
Decision tables retain their usual semantics as described by \citeN{Calvanese2018}.
We briefly recall this semantics.
Each cell $(i, j)$ of a decision table corresponds to a formula $F_{ij}(x)$ in one free variable $x$. 
For instance, a cell \cellcontent{$\leq 50$} corresponds to the formula ``$x \leq 50$''.
A decision table with the \emph{Unique} or \emph{Any} hit policy with rows $R$, input columns $I$ and output columns $O$ is a conjunction of material implications:

\begin{equation} \label{eq:dt}
\bigwedge_{i \in R} \bigg(\bigwedge_{j \in I} F_{ij}(H_j) \Rightarrow \bigwedge_{k \in O} F_{ik}(H_k)\bigg)
\end{equation}

where $H_j$ is the header of column $j$. 
When a decision table is incomplete, it is possible no rows match. In this case, the output given a special \emph{null} value, i.e.,

\begin{equation} \label{eq:null}
\bigg(\bigwedge_{i \in R} \lnot\bigwedge_{j \in I} F_{ij}(H_j)\bigg) \Rightarrow \bigwedge_{k \in O} H_k = \mathit{null}
\end{equation}

In short, a decision table is satisfied when, for each row of which all the input conditions are met, all output conditions are also met. When no rows are applicable, the output is forced to \emph{null} (or the table's default value, if it has one).

For example, in standard DMN (which does not contain function or relation expressions), the table in Figure~\ref{fig:decisiontable} corresponds to the logical formula $(\mathit{Age Of Person} \geq 18 \Rightarrow \mathit{Person Is Adult} = \mathit{Yes}) \land  (\mathit{Age Of Person} < 18 \Rightarrow \mathit{Person Is Adult} = \mathit{No})$.

Data tables are simply a specific case of decision tables.

The semantics of simple constraint tables (without quantification and functions) is also a conjunction of implications, as we described in~\cite{Deryck}. 
The semantics of constraint tables and decision tables differ in the interpretation of incomplete tables: 
when no rows are applicable in constraint tables, its outputs can take any arbitrary value instead of being forced to null (or some default value).

We now extend this semantics to take variables and quantification into account.
Our first step is to define a function that maps cDMN expressions to terms. 
For the most part, this definition corresponds to that of \citeN{Calvanese2018}.

Similarly to Calvanese~et~al., we translate most of the entries $c$ in a cell $(i,j)$ of a table into a formula $F_{ij}(x)$ in one free variable $x$.
For an expression $e$, we denote by $t(e)$ the logical term that corresponds to $e$.
In standard DMN, the only expressions we need to consider are constants and arithmetic expressions built from constants.
In this case, we can simply consider $t(e) = e$.
We will show below how to extend $t$ to the other kinds of expressions in cDMN.
We now define:

\begin{itemize}
	\item If $c$ is of the form ``$\theta e$'' with $\theta$ one of the relational operators $\{\leq, \geq, =, \neq\}$, then $F_{ij}(x)$ is the formula $x ~ \theta ~ t(e)$;
	\item If $c$ is of the form $\mathit{Not}~e$, then $F_{ij}(x)$ is $x \neq t(e)$;
	\item If $c$ is a list $e_1, \ldots, e_n$, then $F_{ij}(x)$ is $x = t(e_1) \lor \ldots \lor x = t(e_n)$.
	As a special case, if $c$ consists of a single expression e, then $F_{ij}(x)$ is $x = t(e)$.
	\item If $c$ is a range, e.g. $[e_1, e_2)$, then $F_{ij}(x)$ is $x \geq t(e_1) \land x < t(e_2)$.
        \item A special case is when $c$ contains ``Yes'' or ``No''. In this case, the header of the column must be an atom $A$ and we translate it into $F_{ij} = A$ or $F_{ij} = \neg A$, respectively.
\end{itemize}

We now extend this transformation to take into account the fact that certain expressions -- which we call {\em variable expressions} -- must be translated to FO variables. 
There are two kinds of variable expressions, as we described in Section~\ref{sub:decision_constraint}. 
We define a mapping $\nu$ that maps each of these two kinds of cDMN variable expressions to a typed \fodot{} variable $x$ of type $T$, which we denote as $x[T]$.
We first define a mapping $\nu_H$ for variable expressions that appear in a header $H$ of a variable introducing column:

\begin{itemize}
	\item The name $T$ of a type is a variable expression.
	We define $\nu_H (T) = x_{T}[T]$, with $x_{T}$ a new variable of type $T$.
	\item An expression $e$ of the form ``$\mathit{Type}$ called $v$'' is a variable expression.
	We define $\nu_H (e) = v[\mathit{Type}]$.
\end{itemize}

We now define a general mapping $\nu$ as follows:
\begin{itemize}
    \item If a variable expression $e$ appears in a header $H$ of a variable introducing column, then $\nu(e) = \nu_H (e)$.
    \item If a variable expression appears elsewhere, then its value is $\nu_H (e)$, where $H$ is the unique variable introducing  header that introduced the variable expression $e$ (see Section~\ref{sub:decision_constraint}).
\end{itemize}

Such a variable expression introduces a new variable in the scope of the table at hand.
Given this function $\nu$, we now define the following mapping $t_\nu(\cdot)$ of cDMN expressions to terms.

\begin{itemize}
	\item The interpretation of a constant, integer or floating point number expression is the constant or number itself.
	That is, for a constant $c$, $t_\nu(c) = c$; similarly, for an integer or floating point number $n$, $t_\nu(n) = n$;
	
	\item For an arithmetic or other expression $e$ of the form $e_1 \theta e_2$ with $\theta \in \{+,-,*,/,<, >, \leq, \geq, =, \neq, \lor, \land \}$, we define $t_\nu(e)~=~t_\nu(e_1)~\theta~t_\nu(e_2)$;
	In other words, the interpretation of such an expression is the operator applied to the interpretation of its sub-expressions.
	
	\item The interpretation of a variable expression is the corresponding variable, i.e., for a variable expression $v$, we define $t_\nu(v) = \nu(v)$.

        \item If $c$ is of the form $\#\mathit{Type}$, then $t_\nu(c) = \#\{\mathit{x}[\mathit{Type}]: \mathit{true}\}$, an \fodot{} aggregate that denotes the number of elements in the type itself.

	\item The interpretation of a function expression is that function applied to the interpretation of each of its arguments.
	For a function expression $F$ of the form ``$\mathit{Function}$ of $\mathit{arg}_1$ and \ldots and $\mathit{arg}_n$'', we define ${t_\nu(F) = \mathit{Function}(t_\nu(\mathit{arg}_1),\ldots, t_\nu(\mathit{arg}_n))}$.
	\item The interpretation of a relation expression is that relation applied to the interpretation of each of its arguments.
	For a relation expression $R$ of the form ``$\mathit{Relation}$ for $\mathit{arg}_1$ and \ldots and $\mathit{arg}_n$'', we define ${t_\nu(X) = \mathit{Relation}(t_\nu(\mathit{arg}_1),\ldots, t_\nu(\mathit{arg}_n))}$.
\end{itemize}

We are now ready to define the semantics of a constraint table.
If $I$ is the set of input columns of the table, $O$ the set of output columns and $V \subseteq I$ the set of variable introducing columns, we define the semantics of the table $T$ as the following formula $\phi_T$:

\begin{equation} \label{eq:ct}
    \bigforall_{l \in V} \nu(H_l): \bigwedge_{i \in R} \bigg(\bigwedge_{j \in I} t_\nu\big(F_{ij}(t_\nu(H_j))\big) \Rightarrow \bigwedge_{k \in O} F_{ik}\big(t_\nu(H_k)\big)\bigg)
\end{equation}

\new{where we quantify over each variable $x$ of type $U$ for which $x[U]$ is the variable $\nu(H_l)$ that corresponds to the variable introducing column $l \in V$.}
In other words, for each tuple of elements of the variables' types, all table rows should be satisfied.
Such a row is satisfied when, if all input conditions are met, all its output conditions are also met.

For example, in Figure~\ref{fig:maxshift}, ${\nu(H_1) = { x[ \mathit{Doctor}] }}$  and ${t_\nu(H_1) = x}$, 
${ \nu(H_2) = { y[\mathit{Day}] }}$ and ${t_\nu(H_2) = y}$, $ {t_\nu(H_3) = \mathit{nb\_shifts\_of}(t_\nu(H_1), t_\nu(H_2)) = \mathit{nb\_shifts\_of}(x, y)}$, which leads to the formula:
\[\forall x[\mathit{Doctor}], y[\mathit{Day}]: \mathit{nb\_shifts\_of}(x, y) \leq 1.\] 
which states that every person $x$ works a maximum of 1 shift for every day $y$.

This semantics generalizes that of regular DMN tables.
Indeed, in regular DMN there are no variables, thus $V = \emptyset$, and only \emph{constant} symbols are allowed, so $F_{ij}(t_\nu(H_j)) = F_{ij}(H_j)$.
As a result, Equation~\ref{eq:ct} simplifies to that in Equation~\ref{eq:dt}.

Decision tables with multiple hit policies have a different semantics. 
We first describe the semantics of \cellcontent{C+}, \cellcontent{C\textless} and \cellcontent{C\textgreater} tables, which are almost identical.
We define the semantics of a \textit{C+} table with one output header $H_k$:

\begin{equation} \label{eq:sumtable}
\bigforall_{w \in W} \nu(H_w):
t_\nu(H_k) = 
\sum_{i \in R}
\mathit{sum}
\big\{ 
\bar{x}:~\bigwedge_{j \in I} F_{ik}\big(t_\nu(H_j)\big): F_{ik}((\bar{x}))
\big\}
\end{equation}
Here, $W \subseteq V$ is the subset of variable introducing columns $V$ of which the variable appears in the output header $t_\nu(H_k)$, $\bar{x}$ are the variables introduced by the remaining variable introducing columns $U = V \setminus W$ (so $\bar{x}$ = $(t_v(H))_{H \in U}$), and $\mathit{sum}\{\bar{x}: \varphi(\bar{x}): F(\bar{x})\}$ denotes the sum of all $F(x)$ for which $\varphi(x)$ holds.

This formula can be explained as follows.
First, when no variables are introduced (i.e., $U = V = W = \emptyset$), this formula sums the output values $F_{ik}$ for each of the rows $i$ that meet the input criteria $\bigwedge_{j \in I}F_{ij}$.
This is precisely the definition of a standard DMN \textit{C+} table.

Second, when variables are introduced in a table, but the output header contains no variables ($W=\emptyset$), it is again assigned a sum of terms.
For each row $i$ and tuple $\bar{x}$ that satisfy $\bigwedge_{j \in I}F_{ij}(\bar{x})$ is satisfied, the value $F_{ik}(\bar{x})$ is included in the sum.

Third, when the output header does contain variables, the table defines the value not of a single constant $H_k$, but of a function $H_k(\nu(\bar{w}))$.
For each appropriate tuple $\bar{a}$, the value of $H_k(\bar{a})$ is defined by the same sum as before.

\begin{figure}
	\centering
        \includegraphics[width=\linewidth]{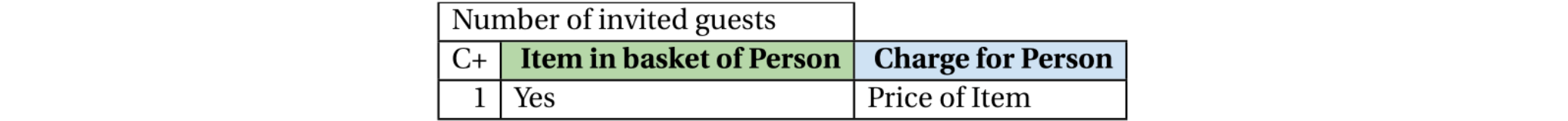}
	\caption[Decision table that determines the charge of a person]{Decision table that determines the charge of a person, based on the contents of their shopping basket.}
	\label{tab:multihit}
\end{figure}

We illustrate this semantics with an example. 
In the decision table shown in Figure~\ref{tab:multihit}: $W=\{\mathit{Person}\}$, 
$\nu(H_w)=p[\mathit{Person}]$, 
$t_\nu(H_k)=Charge(p)$, 
$\bar{x}=y[\mathit{Item}]$, 
$t_\nu(F_{1k}(\bar{x}))~=~\mathit{Price}(y)$ and 
$\bigwedge_{j~\in~I} F_{ij}(t_\nu(H_j))~=~\mathit{InBasket}(y,p)$. 
This results in the logical sentence: 
\[\forall~p[\mathit{Person}]: \mathit{Charge}(p) = \mathit{sum}\{(\mathit{Item}): 
	\mathit{InBasket}(Item, p):
	\mathit{Price}(\mathit{Item})\}\] 

The semantics of \textit{C\textless} and \textit{C\textgreater} tables are defined analogously, where, instead of summing all values, the minimum and maximum value is selected respectively.

Decision tables with a \textit{C\#} hit policy have a slightly different semantics, i.e.,
\begin{equation} \label{eq:counttable}
\bigforall_{w \in W} \nu(H_w):
t_\nu(H_k) = 
\big| \big\{ x \mid \bigexists_{u \in U} \nu(H_u): \bigvee_{i \in R} \big( t_\nu(F_{ik}(x) ) \land \bigwedge_{j \in I} t_\nu(F_{ij}(H_j)) \big)
\big\} \big|
\end{equation}
Here, $U$ and $W$ are defined analogously as in Equation~\ref{eq:sumtable}.

This formula can be explained as follows:
first, when the output header contains no variables ($W = \emptyset$ and $U = V$), the aggregate expression counts for how many $x$'s there exists an assignment of values to the variables $P$ that causes at least one row $i$ of the table to be applicable, in the sense than both its input and output columns are satisfied.
The output header is assigned the size of the set $x$ given that  there exists an expansion of variables for which one of the rules that has $x$ as output fires. 

As before, when the output header does contain variables, for each tuple $\nu(\bar{w})$, the value of $H_k(\nu(\bar{w}))$ is defined in this way.

\begin{figure}
	\centering
        \includegraphics[width=\linewidth]{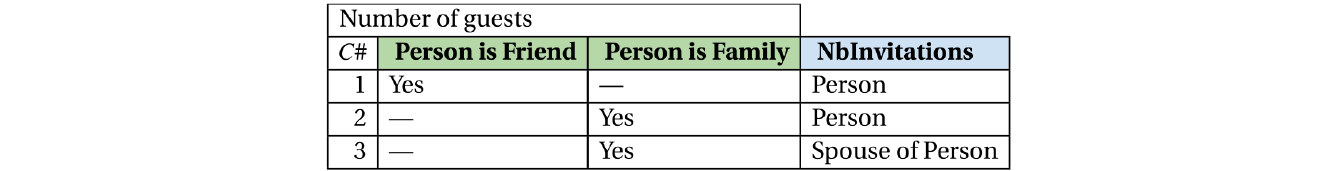}
	\caption{Decision table that counts the number of invited guests.}
	\label{tab:collectcount}
\end{figure}

For instance, in the decision table in Figure~\ref{tab:collectcount}: 
$W=\emptyset$, 
$t_\nu(H_k) = NbInvitations$,
$U=\{Person\}$,
$\nu(H_u)=p[Person]$.
For the first row, $\big( F_{1k}(t_\nu(x) ) \land \bigwedge_{j \in I} F_{1j}(t_\nu(H_j))$ is equivalent to $x=p \land isFriend(p)$.
The second row is defined analogously to the first row. In the third row, $F_{3k}(t_\nu(x) ) \land \bigwedge_{j \in I} F_{3j}(t_\nu(H_j))$ translates to $x=Spouse(p) \land isFamily(p)$.
Consequently, the table in Figure~\ref{tab:collectcount} is logically equivalent to:

\begin{equation}
\begin{aligned}
&NbInvitations = \big|\big\{ x \mid \exists p[Person]: \\
&~~~~~~ x=p \land isFriend(p) \lor \\
&~~~~~~ x=p \land isFamily(p) \lor \\
&~~~~~~ x=Spouse(p) \land isFamily(p) \\
&\big\}\big|
\end{aligned}
\end{equation}

In the table of Figure~\ref{tab:collectcount}, the output header is a constant ($\mathit{NbInvitations}$), therefore no quantification is required.
The value of this constant is calculated as the number of persons that are either friends, family or the spouse of family, while ensuring that duplicate persons (such as friends that are also family) are not counted multiple times.

With this, we have defined the semantics of cDMN.
The goal table that can also be included in a cDMN specification does not contribute to the semantics, but simply tells the cDMN solver what to compute; this can either be a set of possible worlds (one, all, or a specific number of them) or the possible world that minimizes/maximizes a given term.

\section{Implementation} \label{sec:implementation}

Because cDMN is more expressive than DMN, it cannot be handled by existing solvers.
We have therefore implemented a new solver\footnote{https://gitlab.com/EAVISE/cdmn/cdmn-solver}, which we describe in this section.
\old{The solver}\new{It} consists of two parts: an off-the-shelf constraint solver (the IDP system \cite{DeCatBroes2018PLaa}), and a converter from cDMN to IDP input.
In principle, any constraint solver could be used, but we use the IDP system because it directly supports \fodot{}.
The input of the system is a cDMN model created in a spreadsheet in the \texttt{.xlsx} format.
Such a spreadsheet allows for straightforward creation of cDMN tables, and can show a good overview over the entire model.
The cDMN to IDP converter is written in Python, and works in a two-step process.

First, the converter interprets all tables in a spreadsheet, and converts them into Python objects.
For example, the converter parses all the glossary tables and converts them into a single \texttt{Glossary} object, which then creates \texttt{Type} and \texttt{Predicate/Function} objects.
The constraint and decision tables are then evaluated individually.
A lex/yacc parser inspects each cell and parses it as a cDMN expression, such as ``\textit{Function} of $\mathit{arg}_{1}$ and $\mathit{arg}_{2}$''.
Such cDMN expressions are then interpreted using the constructed \textit{Glossary} object, and transformed into an \fodot{} expression, e.g. ``$\mathit{Function}(\mathit{arg}_{1}, \mathit{arg}_{2}$)''.
\new{For example, the expression ``nb shift of Golgi on d2'' is transformed into ``\textit{nb\_shift\_of\_Doctor\_on\_Day}$($\textit{Golgi}, \textit{d2}$)$''.}
Each cDMN table is then converted to an \fodot{} formula, as described in Section \ref{ss:semantics}.

The created Python objects are then converted into IDP blocks.
The knowledge in the IDP system is structured in three such blocks: the \textit{vocabulary}, the \textit{structure} and the \textit{theory}.
On top of these, there are also the \textit{main} and \textit{term} blocks.
The \textit{main} block is used to specify the logical inference method that should be applied.
cDMN makes use of two of these inference methods: \textit{model expansion} (find an expansion $S' \supseteq S$ of a structure $S$ for part of the vocabulary of a theory $T$ such that $S' \models T$) and \textit{optimization} (find the model expansion of $S$ w.r.t. $T$ that minimizes/maximizes a given a numerical term).
The \textit{term} block is used to specify the optimization term when optimizing.

\begin{figure}
    \includegraphics[height=\linewidth, angle=270]{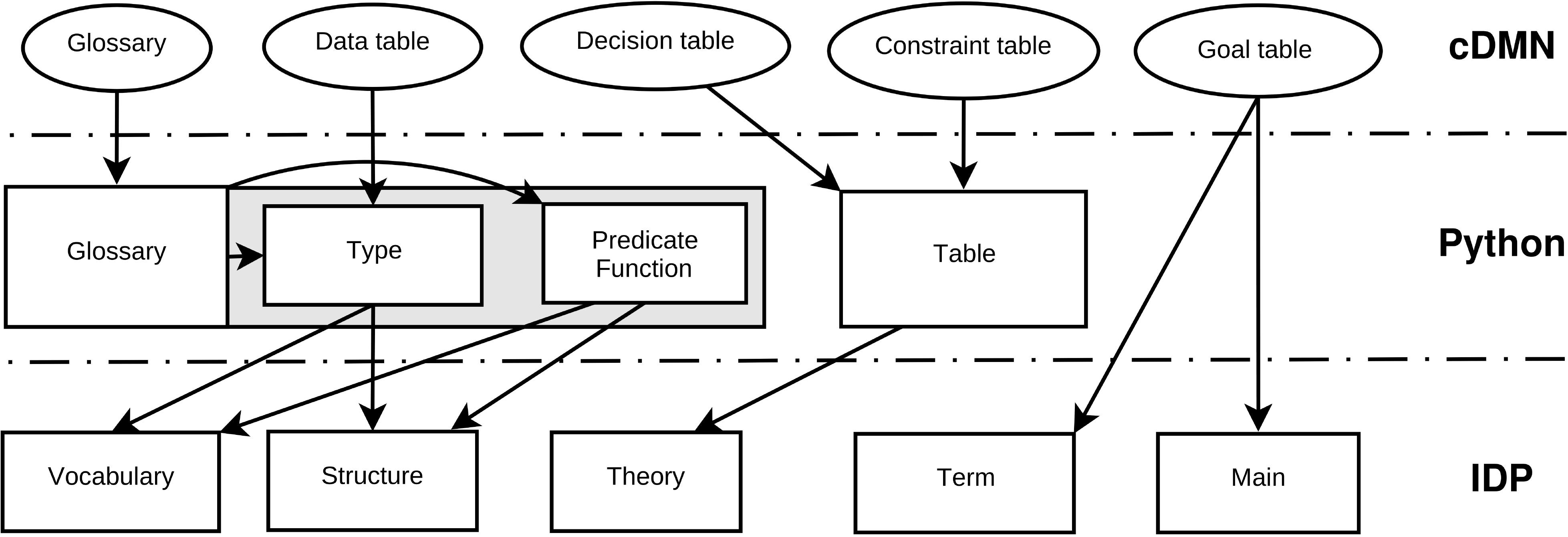}

    \caption{An overview of the inner workings of the cDMN solver}
    \label{fig:converter}
\end{figure}

An overview summarizing all the relations between cDMN tables, Python objects and IDP blocks can be found in Figure~\ref{fig:converter}.
More detailed information about this conversion can be found in the cDMN documentation\footnote{www.cdmn.be}, along with an explanation of the usage of the solver and concrete examples of cDMN implementations.

Besides cDMN tables, the solver also supports most standard DMN tables \new{and constructs.}
\old{Only tables with the \cellcontent{C} hit policy (which collect the outputs of all matching rows in a list, rather than, e.g., summing them as the \cellcontent{C+} hit policy does) or tables containing boxed expressions (a more complicated form of the FEEL language, which is generally considered to be less usable by domain experts) are not supported.}
\new{More specifically, it supports tables with the \cellcontent{U}, \cellcontent{A}, \cellcontent{F}, \cellcontent{C+}, \cellcontent{C$>$} and \cellcontent{C$<$} hit policy, and the full S-FEEL language.
While there is currently no support for the \cellcontent{C}-tables, which collect the outputs of all matching rows in a list, it is possible to use a cDMN relation to emulate such a table's behaviour.}

\new{Standard} DMN specifications can be supplied in the form of a spreadsheet (as for cDMN), but also in the XML format specified in the DMN standard.
As such, the solver can also be used as a drop-in replacement for standard DMN tools, allowing for a more flexible usage of the knowledge in a DMN model \cite{VandeveldeSimon}.

Another feature of the solver is the ability to link to the IDP-based interface \textit{Interactive Consultant} \cite{IC}, which is a user-friendly interface for interactively solving configuration problems.
It shows users the consequences of their choices and provides explanations for these consequences.
Thus by combining cDMN and the Interactive Consultant interface, a KB can be both constructed and interacted with in a user-friendly manner.

\section{Results and discussion} \label{sec:results}

In this section we first look at three of the DM Community challenges, each showcasing a feature of cDMN.
For each challenge, we qualitatively compare the DMN implementations from the DM Community website with our own implementation in cDMN.
Afterwards, we compare all challenges on size and quality.
\new{We end our discussion with a section on the integration of cDMN in business processes.}

\subsection{Constraint tables}
\begin{figure}
    \includegraphics[width=\linewidth]{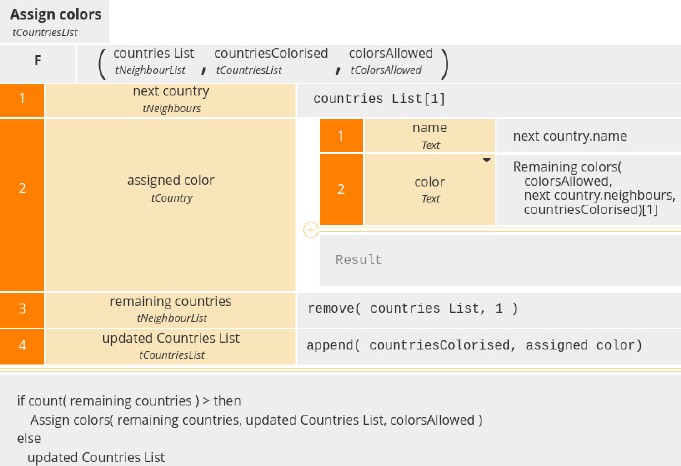}

    \caption{An extract of the map coloring solution in standard DMN with FEEL}
    \label{fig:FEELmap}
\end{figure}
Constraint tables allow cDMN to model constraint satisfaction problems in a straightforward way.
For example, in \chalname{Map Colouring}, a map of six European countries  must be colored in such a way that no neighbouring countries share the same color.
For this challenge, a pure DMN implementation was submitted, of which Figure \ref{fig:FEELmap} shows an extract.
The implementation uses complicated FEEL statements to solve the challenge. 
While these statements are DMN-compliant, they are nearly impossible for a business user to write without help.
In cDMN, we can use a single straightforward constraint table to solve this problem, as shown earlier in Figure \ref{tab:map}.
Together with the glossary and a data table (Figure \ref{tab:datatable_example}), this forms a complete yet simple cDMN implementation.

\subsection{Quantification}
Quantification proves useful in the \chalname{Monkey Business} challenge.
In this challenge, we want to know for four monkeys what their favourite fruit and their favourite resting place is, based on some information.
There are two DMN-like submissions for this challenge: one using Corticon, and one using OpenRules.

\begin{figure}
    \centering
    \begin{subfigure}[t]{\textwidth}
        \centering\includegraphics[width=10cm]{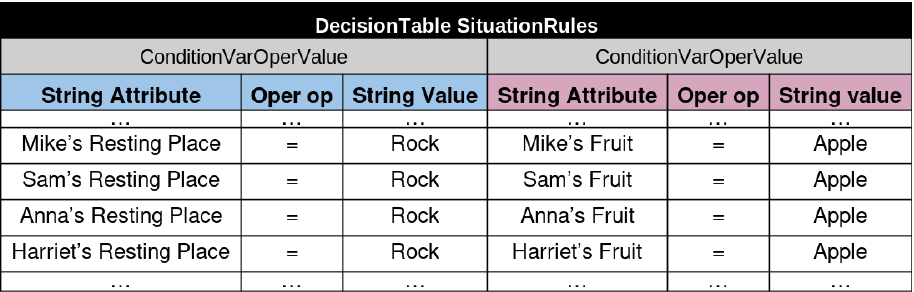}
        \caption{Open Rules}
        \label{fig:ormonkey}
    \end{subfigure}
    
    \begin{subfigure}[c]{\textwidth}
        \centering
        \includegraphics[width=\linewidth]{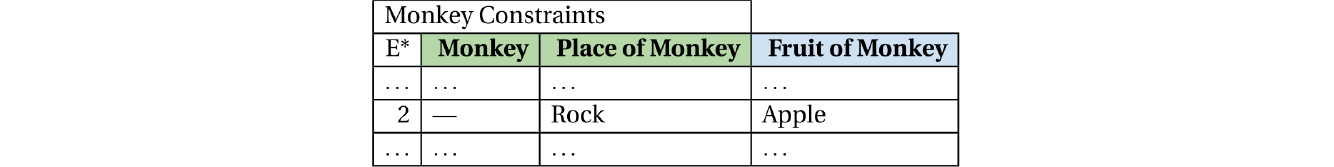}
        \caption{cDMN}
        \label{tab:monkey_business2}
    \end{subfigure}
    \caption{An extract of \chalname{Monkey Business} implementation in (a) OpenRules and (b) cDMN, specifying \specification{The monkey who sits on the rock is eating the apple}.}
\end{figure}

One of the pieces of information is: \specification{The monkey who sat on the rock ate the apple.}
The OpenRules implementation has a table with a row for each monkey, which states that if this monkey's resting place was a rock, their fruit was an apple (Figure \ref{fig:ormonkey}).
In other words, for $n$ monkeys, the OpenRules implementation of this rule requires $n$ lines.
Because of quantification, cDMN requires only one row, regardless of how many monkeys there are (Figure \ref{tab:monkey_business2}).
The Corticon implementation also uses a similar quantification for this rule.

\begin{figure}
    \begin{subfigure}[c]{\linewidth}
        \centering
        \includegraphics[width=8cm]{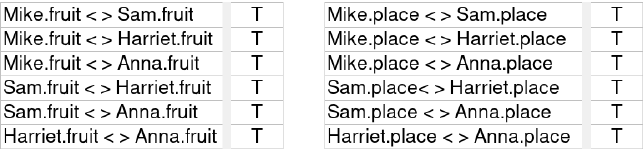}
        \caption{Corticon}
        \label{fig:cormonkey}
    \end{subfigure}
    
    \begin{subfigure}[c]{\linewidth}
        \centering
        \includegraphics[width=\linewidth]{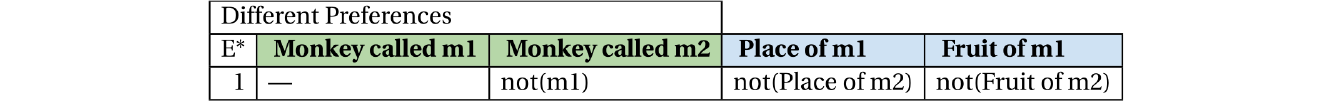}
         \caption{cDMN}
         \label{tab:monkey_business1}
    \end{subfigure}
    \caption{An extract of the \chalname{Monkey Business} implementation in (a) Corticon and (b) cDMN, defining that no monkeys share fruit and no monkeys share the same place.}
\end{figure}

Another rule states that no two monkeys can have the same resting place or fruit.
In both the Corticon and OpenRules implementations, this is handled by two tables with a row for each pair of monkeys.
The Corticon tables are shown in Figure \ref{fig:cormonkey}.
Each row either states that two monkeys have different fruit, or that they have different place.
Therefore, $n$ monkeys require $ \frac{n\times(n-1)}{2} $ rows.
By contrast, the cDMN implementation in Figure \ref{tab:monkey_business1} requires only a single row to express the same.

We conclude that of all the solutions that were submitted to the DM Community, only the cDMN solution has quantification powerful enough to represent the constraints of this puzzle in a way that is independent of the size of the problem instance\new{.}\old{, while still remaining readable and user-friendly for non-experts.} 

\subsection{Optimization}
\begin{figure}
    \includegraphics[width=\linewidth]{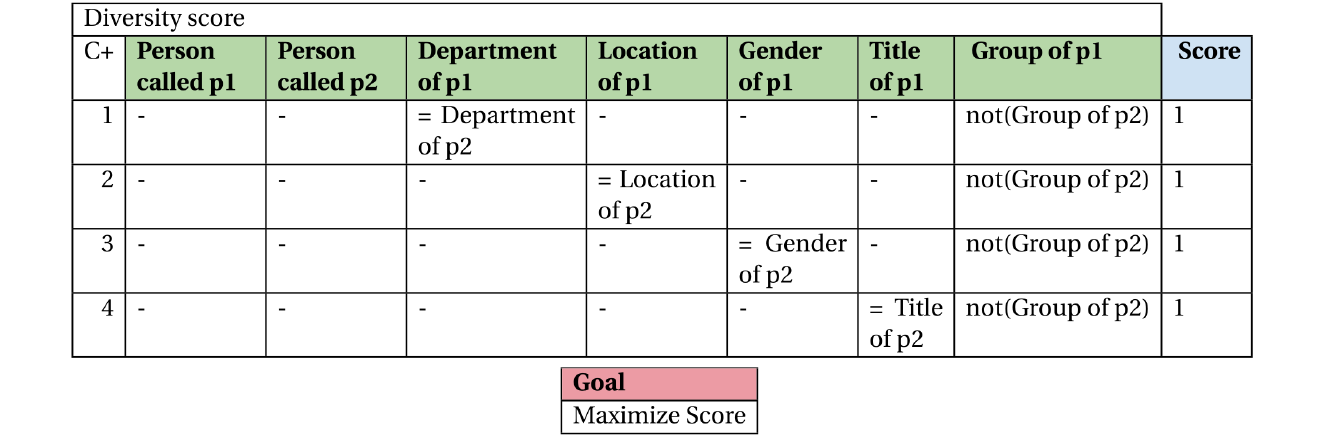}
    \caption{The decision tables and constraint table for \chalname{Balanced Assignment}.}
    \label{tab:balanced_assignment}
\end{figure}

In the \chalname{Balanced Assignment} challenge, 210 employees need to be divided into 12 groups, so that every group is as diverse as possible.
The department, location, gender and title of each employee is known.
This is quite a complex problem to handle in DMN.
As such, of the four submitted solutions, only one was DMN-like: an OpenRules implementation, using external CP/LP solvers.
The logic for these external solvers is written in Java.
Although the code is fairly compact, it cannot be written without prior programming knowledge.
The optimization \old{features of}\new{support in} cDMN\old{,} allows us to \old{solve} \new{represent} the problem with two decision tables and one constraint table.
The table \textit{Diversity score}, shown in Figure \ref{tab:balanced_assignment}, adds 1 to the total diversity score if two similar people are in a different group. Maximizing this score then results in the most diverse groups.

\new{While it is possible to model this problem using the cDMN notation, the internal engine in the cDMN solver can not find a solution in reasonable time due to the large problem size.
However, our solver is only a reference implementation; it might be possible to create other solvers for cDMN that would be capable of solving this problem.
}

\subsection{Overview of all challenges}

Of the 24 challenges we considered, cDMN is capable of successfully modelling 22.
In comparison, there were 12 OpenRules implementations and 12 Corticon implementations submitted.
Note that we cannot rule out that OpenRules and Corticon might be capable of modelling more challenges than those for which a solution was submitted.

To compare cDMN to other approaches, we focus on two aspects. First, we quantitatively measure the size of the solution.\old{ quantity (how big are the solutions?) and quality (how readable and how scalable are they?).}
\old{The size of implementations} \new{This} was measured by counting the number of cells used in all the decision and constraint tables. 
We exclude meta information (such as the cDMN glossary) and the specification of a concrete problem instance (such as the cDMN data tables), because the ways in which different solvers handle this are too diverse to allow meaningful comparison.
Table \ref{table:cellcount} shows that cDMN and Corticon alternate between having the fewest cells, and that OpenRules usually has the most.
In general, OpenRules implementations require many cells because each cell is very simple.
For instance, even an ``='' operator is its own cell.
The Corticon implementations, on the other hand, contain more complex cells, rendering them more compact.

\new{
Second, we also qualitatively assess the readability and scalability of the solutions.
The motivation for this is that model size, as we have defined it above, does not tell the whole story.
Indeed, using very complex expressions might lead to small tables, that are nevertheless hard to figure out.
}

\begin{table}
    \centering
    \caption{Comparison of the number of cells used per implementation. Lowest number of cells per challenge in grey. Other implementations: 1. FEEL, 2. Blueriq, 3. Trisotech, 4. DMN}
    \label{table:cellcount}
    \begin{tabular}{c c c c c}
        \hline \hline
        & cDMN & Corticon & OpenRules & Others \\
        \hline
        \LCC%
        & \graycell  & & \\
        \Color{0 0 0} Who Killed A.? & 53 & 54 & 176 & / \\
        \ECC
        \LCC%
        & & \graycell & \\
        \resetcolor Change Making & 26 & 14 & / & / \\
        \ECC
        \LCC%
        & \graycell & & \\
        \resetcolor A Good Burger & 35 & 20 & 95 & $76^1$ \\
        \ECC
        \LCC%
        & & \graycell & \\
        \resetcolor Define Dupl. & 20 & 19 & 21 & / \\
        \ECC
        \LCC%
        & \graycell & & \\
        \resetcolor Coll. of Cars & 26 & 45 & / & $48^1$ \\
        Monkey Business & 47 & 64 & 150 & / \\
        \ECC
        \LCC%
        & & & & \graycell \\
        \resetcolor Vacation Days & 38 & 32 & 31 & $14^2$ \\
        \ECC
        \LCC%
        & & \graycell \\
        \resetcolor Family Riddle & 76 & 22 & / & / \\
        \ECC
        \LCC%
        & \graycell \\
        \resetcolor Cust. Greeting & 88 & / & 205 & / \\
        Online Dating & 45 & 78 & / & / \\
        \ECC
        \LCC%
        & & \graycell \\
        \resetcolor Class. Employees & 36 & 21 & 70 & $34^3$ \\
        \ECC
        \LCC%
        & \graycell \\
        \resetcolor Reinder Order & 14 & 64 & 111 & $370^4$ \\
        Zoo, Buses, Kids & 24 & / & 43 & / \\
        \ECC
        \LCC%
                         & & & \graycell \\
        \resetcolor Balanced Assign. & 55 & / & 30 & / \\
        Vac. Days Adv. & 124 & / & 97 & / \\
        \ECC
        \LCC%
        & \graycell \\
        \resetcolor Map Coloring & 21 & / & / & $34^4$ \\
        Map Color Viol. & 21 & / & / & / \\
        Crack The Code & 48 & / & / & / \\
        Numerical Haiku & 41 & / & / & \\
        Nim Rules & 22 & / & 61 & / \\
        Doctor Planning & 102 & / & / & / \\
        Calculator & 33 & /  & /  & /  \\
        \ECC
        \hline \hline
    \end{tabular}
\end{table}

\begin{figure}
    \centering
    \begin{subfigure}[c]{\linewidth}
        \centering\includegraphics[width=8cm]{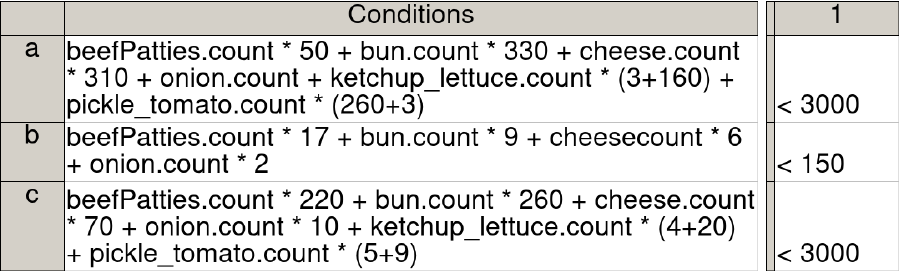}
        \caption{Corticon}
        \label{fig:hamburger_comparison}
    \end{subfigure}
    
    \begin{subfigure}[c]{\linewidth}
        \centering
        \includegraphics[width=\linewidth]{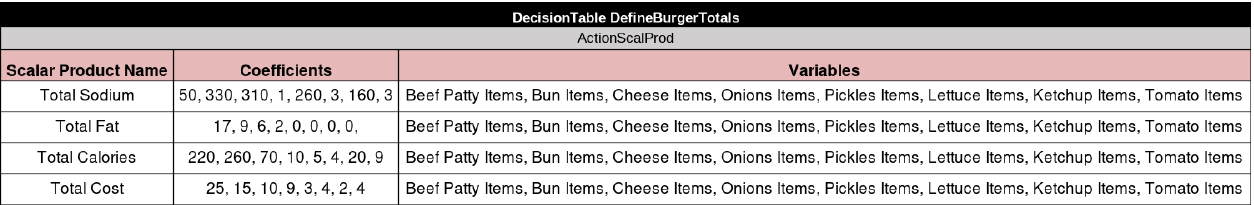}
        \caption{OpenRules}
        \label{fig:hamburger_openrules}
    \end{subfigure}
    
    \begin{subfigure}[c]{\linewidth}
        \centering
        \centering\includegraphics[width=\linewidth]{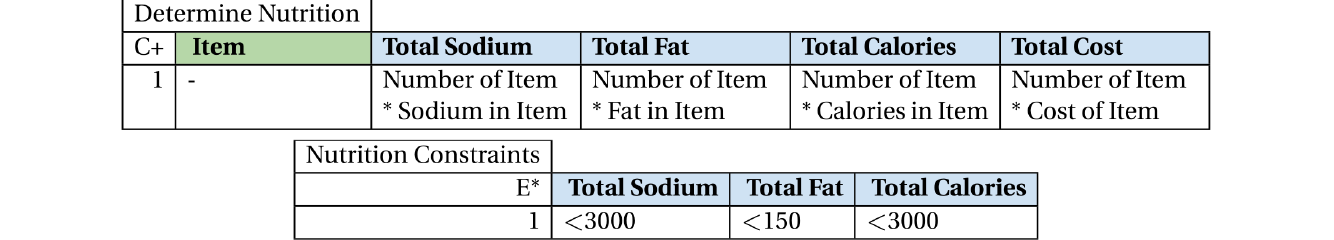}
        \caption{cDMN}
        \label{tab:hamburger_comparison}
    \end{subfigure}
    \caption{Calculating the food properties of a burger in Corticon, OpenRules and cDMN.}
\end{figure}

\old{The other side of the coin, is} \new{In general, we find} that OpenRules implementations are usually easier to read than their Corticon counterparts.
An example comparison between cDMN and Corticon can be seen in Figure \ref{fig:hamburger_comparison} and \ref{tab:hamburger_comparison}. 
Each figure shows a snippet of their \chalname{Make a Good Burger} implementation, in which the food properties of a burger are calculated.
While the Corticon implementation is more compact, it is less interpretable, less maintainable and dependent on domain size.
If the user wants to add an ingredient to the burger, complex cells need to be changed.
In cDMN, we introduce a type \cdmnvar{Ingredient}, a number of functions such as \cdmnvar{Amount of Ingredient} and \cdmnvar{Fat in Ingredient}, and calculate the constant \cdmnvar{Total Fat} as the product of the fat in a specific ingredient and the amount of that ingredient used.
This enables the user to simply add new ingredients or change the amount of nutrition values in the data table, without having to change the model.
The OpenRules implementation (Figure~\ref{fig:hamburger_openrules}) is fairly readable and modular too, but, it requires a custom scalar product decision table.

\begin{figure}
    \begin{subfigure}[c]{\linewidth}
    \centering
    \includegraphics[width=8cm]{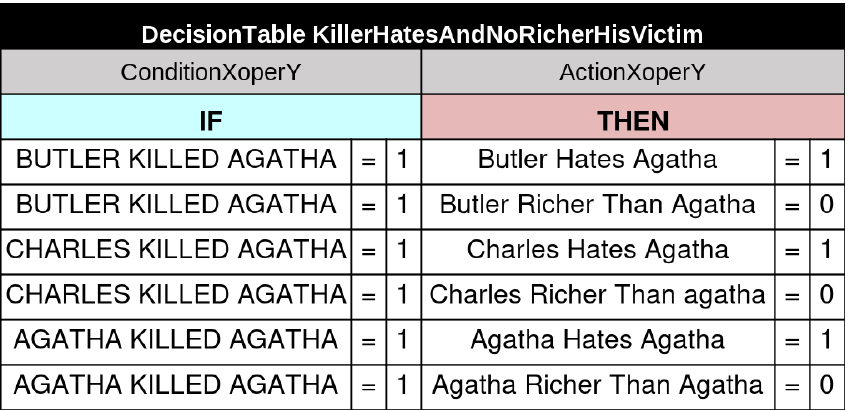}
    \caption{OpenRules}
    \label{fig:agatha_comparison}
    \end{subfigure}
    
    \begin{subfigure}[c]{\linewidth}
        \centering
        \includegraphics[width=\linewidth]{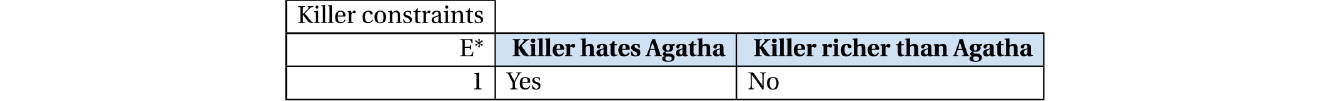}
        \caption{cDMN}
        \label{tab:agatha_comparison}
    \end{subfigure}
    \caption{Implementation of \specification{A killer always hates and is no richer than their victim} in OpenRules and cDMN}
\end{figure}

Another comparison between cDMN and OpenRules can be found in Figure \ref{fig:agatha_comparison} and \ref{tab:agatha_comparison}.
Here we show \old{a part} \new{a snippet} of the \chalname{Who Killed Agatha?} challenge.
Both show a translation of the following rule: \specification{A killer always hates, and is not richer than, his victim.}
By using constraints and a constant (\cdmnvar{Killer}), cDMN allows us to form a \old{more readable and} more scalable table.
\new{Indeed, i}\old{I}f the police ever find a fourth suspect, they can easily add the person to the data table without needing to change anything else.

In Section \ref{sec:chaloverview}, we identified four relevant problem properties.
We now suggest that each property is tackled more easily by one or more of the additions cDMN proposes.

\textbf{Aggregates needed}
Figure \ref{tab:hamburger_comparison} shows how aggregates are both more readable and scalable when using quantification.
Moreover, cDMN allows the use of aggregates for more complex operations such as optimization or defining constraints.

\textbf{Constraints}
Constraints can be conveniently modelled by constraint tables, such as the constraints in Figure \ref{tab:agatha_comparison}, which state that the killer hates Agatha, but is no richer than her.
The addition of constraint tables allows for an obvious translation from the rule in natural language to the table.

\textbf{Universal quantification}
Problems which contain universal quantification can be compactly represented, as can be seen in Figure \ref{fig:maxshift}. This table states that no doctor works more than one shift per day.

\textbf{Optimization}
Because cDMN directly supports optimization, problems containing this property are easily modelled.
Furthermore, by the addition of more complex data types, optimization terms can be defined in a more flexible manner.
An example can be found in \chalname{Balanced Assignment} in Figure \ref{tab:balanced_assignment}.
A summary of each problem property and its cDMN answer can be found in Table \ref{table:propsol}.

\begin{table}
    \centering
    \caption{Comparison between the problem properties and their cDMN answers.}
    \label{table:propsol}
    \begin{tabular}{l l}
        \hline \hline
        Property & cDMN answer \\
        \hline
        Aggregates needed & Quantification, expressive data \\
        Constraints & Constraint tables, quantification, expressive data \\
        Universal quantification & Quantification \\
        Optimization & Optimization, expressive data \\
        \hline
    \end{tabular}
\end{table}

\subsection{Process integration}

\begin{figure}
    \includegraphics[width=\linewidth]{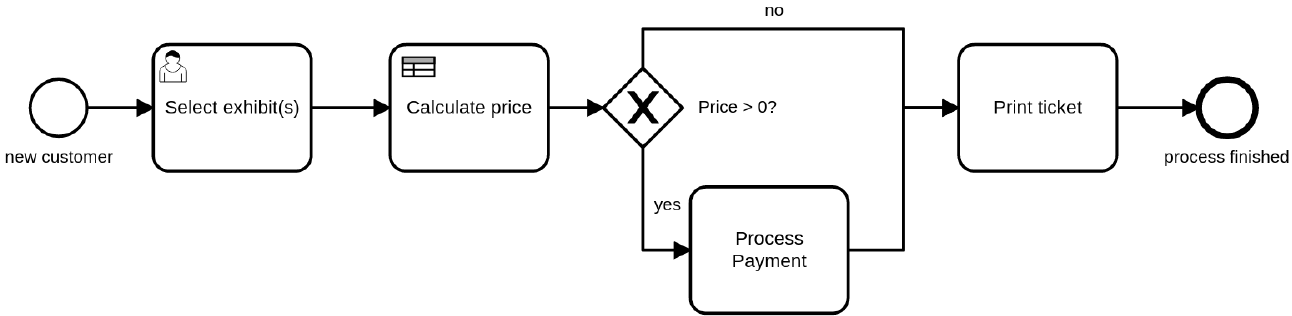}
    \caption{Example of a BPMN model with DMN}
    \label{fig:bpmn}
\end{figure}

\begin{figure}
    \centering
    \includegraphics[width=1\linewidth]{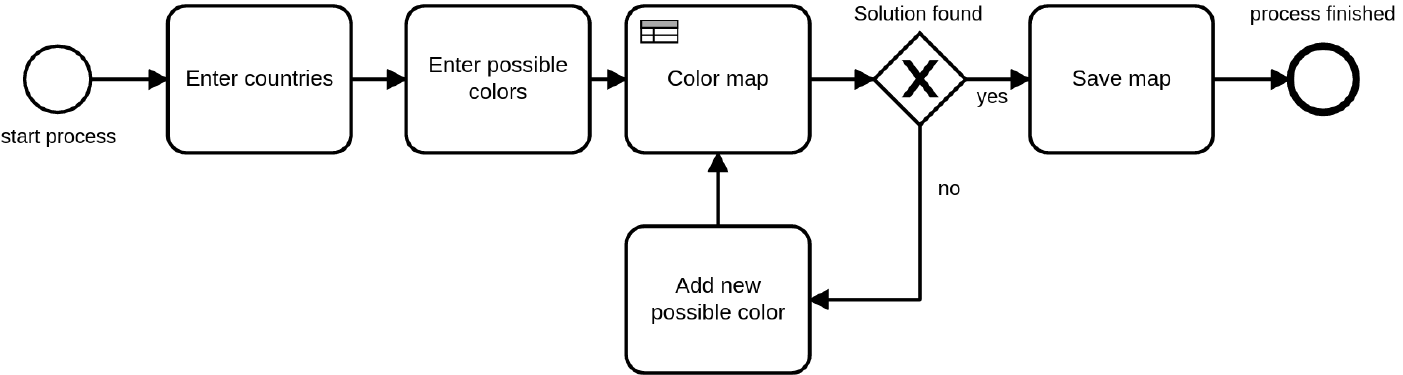}
    \caption{Example of a BPMN model with cDMN}
    \label{fig:cbpmn}
\end{figure}

\new{
DMN models are often integrated into a larger business process model~\cite{Hasic2018,Bazhenova2019}.
Such a business process model consists of a sequence of steps that describe how to execute a specific process, such as for example the steps required for verifying a customer's eligibility for a bank loan.
The Business Process Model and Notation is a standard published by the OMG group for this purpose.

The integration of DMN into BPMN is motivated by the \textit{separation of concerns} paradigm~\cite{sepconcerns}, in which the decision logic is separated from the process, to increase readability and maintainability of the overall process model.
If a DMN model is present in a BPMN model, it can be used to dictate the flow of the process using a so-called \textit{gateway}, depicted by a diamond.
For example, the BPMN model in Figure~\ref{fig:bpmn} describes the flow for buying a ticket to a museum.
After a visitor has selected the exhibits they want to visit, the price of the ticket is determined by the decision model shown in Figure~\ref{fig:drd}.
The output of the decision model then dictates whether a payment is required, or if the ticket can be printed directly (in the event that only free exhibits were selected).

In principle, cDMN models could also be used in a BPMN model to direct the flow of a process.
When a DMN model is used in BPMN, the process is always directed based on  the value of the top level variable (such as $\mathit{Price} > 0$ or $\mathit{Eligible} = \mathit{Yes}$) of the DMN model.
By contrast, the integration of cDMN also allows for other criteria.
For example, the model in Figure~\ref{fig:cbpmn} describes the process of coloring a map of countries, based on a list of countries and a list of possible colors.
Here, we have added a gateway that verifies if a suitable solution was found.
If none was found (because too few colors were supplied), more colors are added until a solution becomes possible.
In other words, the direction of the process is based on whether or not a satisfying solution for the cDMN constraints could be found.
Other examples of possible gateway criteria are verifying if at least $n$ solutions exist, if a solution with a value for variable $x$ greater than 5 exists, what the maximum value of variable $y$ is, and more.


} 


\section{Conclusions} \label{sec:conclusion}

This paper presents an extension to DMN, which \new{aims at solving} \old{is able to solve} complex problems while maintaining  DMN's level of readability.
This extension, which we call cDMN, adds constraint modelling, more expressive data representations (such  as types and functions) and quantification.

Constraint modelling allows a user to define a solution space instead of a single solution.
The cDMN solver can generate a desired number of models, or generate the model which optimizes the value of a specific term.
Unlike DMN, which only knows constants, cDMN also supports the use of functions and predicates, which allow for more flexible representations.
Together with quantification, this allows tables to be constructed in a compact and straightforward manner, while being independent of the size of the problem.
This improves \old{readability,} maintainability and scalability of tables.

\old{
The cDRD gives a user-friendly overview of the structure of a cDMN model, allowing a user to quickly see what terms define a variable, and what terms constrain a variable.
This further improves traceability of decisions.
}

By comparing our cDMN implementations to the implementations of other state-of-the-art DMN-like solvers, we can conclude that cDMN succeeds in increasing the expressiveness \new{of DMN.}\old{while retaining the simplicity of standard DMN}
Moreover, our qualitative analysis of these examples suggest that the cDMN representations are indeed typically quite readable and maintainable.
In future work, we plan to investigate this in a more detailed and quantifiable way, and to compare the user-friendliness and complexity of cDMN to that of DMN itself.


\old{In future work\new{,} it would be interesting to further extend} \new{Other future work consists of possibly extending} the cDMN notation to be able to represent disjunctions in the output of a constraint table, existential quantification, quantification in output columns and increase compactness of the created models.
\old{Furthermore} \new{Additionally}, \new{we are planning on testing} \old{it would be interesting to test} this notation in a number of \old{additional} \new{real-life} use-cases to verify its applicability in a multitude of domains.
\new{The insights gained during the implementation of these use cases, will allow us to define a graph-based representation of cDMN models, akin to the DRD for DMN.}

\textbf{Competing interests:} The authors declare none.

\bibliographystyle{acmtrans}
\bibliography{biblio.bib}
\end{document}